\documentclass[conference]{IEEEtran}
\ifCLASSINFOpdf
 \usepackage[pdftex]{graphicx}
\usepackage{xcolor}
\usepackage{hyperref}      
\hypersetup{
    colorlinks = true,                    
    citecolor = {blue},
    linkcolor = {purple},
           }

\else
\fi
\begin{document}
\title{A generic framework for task selection driven by synthetic emotions}

\author{\IEEEauthorblockN{Claudius Gros}
\IEEEauthorblockA{Institute for theoretical physics\\
Goethe University Frankfurt\\
Frankfurt a.M., Germany\\
Email: http://itp.uni-frankfurt.de/~gros}
}

\maketitle
\begin{abstract}
Given a certain complexity level, humanized agents
may select from a wide range of possible tasks, with 
each activity corresponding to a transient goal.
In general there will
be no overarching credit assignment scheme allowing
to compare available options with respect to expected
utilities. For this situation we propose a task selection 
framework that is based on time allocation via emotional 
stationarity (TAES). Emotions are argued to correspond
to abstract criteria, such as satisfaction, challenge and 
boredom, along which activities that have been carried
out can be evaluated. The resulting timeline of experienced
emotions is then compared with the `character' of the agent, 
which is defined in terms of a preferred distribution of 
emotional states. The long-term goal of the agent, to align 
experience with character, is achieved by optimizing the 
frequency for selecting the individual tasks. Upon optimization,
the statistics of emotion experience becomes stationary.
\end{abstract}

\IEEEpeerreviewmaketitle
\section{Introduction}

A wide range of specific tasks can be handled efficiently
by present day robots and machine learning algorithm.
Playing a game of Go is a typical example
\cite{silver2017masteringGo}, with CT-scan 
classification \cite{gao2017classification} and 
micro-helicopter piloting \cite{kumar2017opportunities}
being two other tasks that are accomplished with 
comparative modest computational resources. As a
next step one may consider agents that have not 
just a one, but a range of capabilities. like being 
able to play several distinct board games. Classically, 
as in most today's applications, a human operator 
interacting with the program decides which of the 
agents functionality he or she wants to access.
Humanized agents may however be envisioned to function
without direct supervision, also with respect to
task selection \cite{malfaz2011biologically}. 
Autonomous agents that are endorsed with the capability 
to connect on their own to either a Go or a chess server
could be matched f.i.\ with either a human opponent, 
upon entering the queuing system, or with another 
board-playing program. The problem is then how to 
allocate time \cite{gros2013emotional}, namely how 
to decide which type of game to play.

\begin{figure}[!t]
\centering
\includegraphics[width=0.45\textwidth]{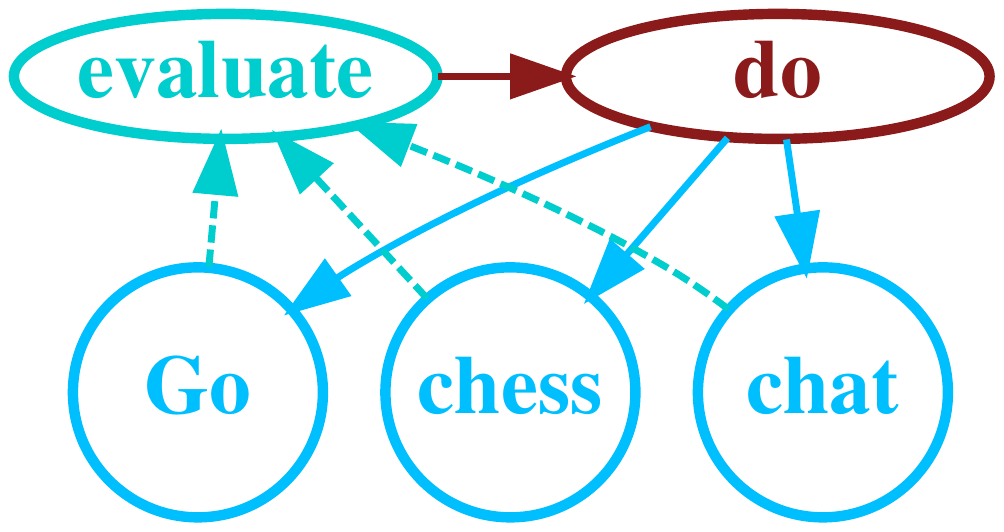}
\caption{Illustration of a general time-allocation framework.
The different options, here to play Go, to play chess and to chat,
are evaluated once selected, with the evaluation results feeding 
back into the decision what to do next.}
\label{fig_frameworkGeneral}
\end{figure}

Deciding what to do is a cornerstone of human 
activities, which implies that frameworks for 
multi-task situations deserve attention. A
possible route is to define and to maximize an
overarching objective function, which could be,
for example, to play alternatively Go or chess 
in order to improve the respective levels of 
expertise, as measured, f.i., by the respective
win rates. This example suggest that time-allocation 
frameworks need two components, see
Fig.~\ref{fig_frameworkGeneral}:
\begin{itemize}
\item A set of criteria characterizing tasks that have
      been executed, the experience of the agent.
\item A set of rules for task switching that are based
      fully or in part on experiences.
\end{itemize}
Implementations may distinguish further between
dedicated and humanized settings. For a specific
dedicated application appropriate hand-picked 
sets of evaluation criteria and switching rules 
can be selected. Within this approach the resulting 
overall behavior can be predicted and controlled to
a fair extend. Being hand-crafted, the disadvantage 
is that extensions and transfer to other domains 
demand in general extensive reworks.

Consider an agent with two initial abilities,
to play Go and chess via internet servers. The
respective winning rates could be taken in this 
case as appropriate evaluation criteria, other 
may be a challenge (close games) and boredom 
(games lasting forever). As an extension, the
agent is provided with a connection to
a chat room, where the task is to answer 
questions. Humans would sent in chess board 
positions and the program provide in return 
the appropriate analysis, e.g.\ in terms of possible
moves and winning probabilities. The program
has then three possibilities, as shown in
Fig.~\ref{fig_frameworkGeneral}, to play Go or chess,
and to connect to the chat room, with the third
task differing qualitatively from the first two.
Time allocation frameworks designed specifically 
for the first two options, to play Go or chess,
would most probably cease to work when the chat room
is added, f.i.\ because winning ratios are not
suitable for characterizing a chat session. Here 
we argue that a characteristic trait of 
humanized computing is universality, which 
translates in the context of time allocation 
frameworks to the demand that extensions to new 
domains should be a minor effort.

Of particular interest to humanized time allocation 
frameworks is emotional control, which is known
to guide human decision making. Starting with an
overview, we will discuss first the computational
and neurobiological role of mammalian emotions,
stressing that algorithmic implementations need
to reproduce functionalities and not qualia
like fear and joy. A concrete implementation
based on the stationarity principle is then
presented in a second step. Synthetic emotions 
correspond in this framework to a combination of
abstract evaluation criteria and motivational
drives that are derived from the objective to 
achieve a predefined time-averaged distribution of
emotional activities, the `character' of the agent.

An alternative to the here explored route to
multi-task problems is multi-objective 
optimization \cite{deb2014multi}, a setting
in which distinct objectives dispose of 
individual utility function that need to be
optimized while respecting overall resource
limitation, like the availability of time.
We focus here on emotional control schemes, noting
that emotional control and multi-objective
optimization are not mutually exclusive.

\section{Computational role of emotions\label{CompEmo}}

Emotions have emerged in the last decades as 
indispensable preconditions for higher cognition 
\cite{panksepp2004affective,gros2010cognition},
with the reason being that the core task of emotional 
response is not direct causation of the type 
``fleeing when afraid'', akin to a reflex, but the 
induction of anticipation, reflection and cognitive 
feedback \cite{baumeister2007emotion}. In general,
being afraid will not result in a direct behavioral 
response, but in the allocation of cognitive resources 
to the danger at hand.

The interrelation between emotion and cognition is
two-faced. Emotions prime cognitive processes \cite{beeler2014kinder},
being controlled in return by cognition \cite{ochsner2005cognitive}. 
The latter capability, to regulate emotions \cite{inzlicht2015emotional}, 
f.i.\ when restraining one's desire for unhealthy food,
is so pronounced that it can be regarded to be a defining 
characteristics our species \cite{cutuli2014cognitive}.
With regard to synthetic emotions, it is important to note 
that the cogno-emotional feedback loop present in our brain
implies that emotional imprints are induced whenever
cognitive capabilities are used to pursue a given goal, 
such as playing and winning a game of Go \cite{miller2018happily}. 

On a neuronal level one may argue \cite{pessoa2008relationship},
that the classical distinction between affective and 
cognitive brain regions is misleading \cite{pessoa2019embracing}.
Behavior should be viewed instead as a complex cogno-emotional 
process that is based on dynamic coalitions of brain 
areas \cite{pessoa2018understanding}, and not on the activation 
of a specific structure, such as the amygdala \cite{phelps2006emotion}. 
This statement holds for the neural representations of the 
cognitive activity patters regulating emotional reactions, 
which are not localized in specific areas, but distributed 
within temporal, lateral frontal and parietal 
regions \cite{morawetz2016neural}.

The mutual interrelation of cognitive and emotional brain 
states suggests a corresponding dual basis for decision 
making \cite{lerner2015emotion}. Alternative choices are
analyzed using logical reasoning, with the outcome being 
encoded affectively \cite{reimann2010somatic}.
Here we use `evaluation criteria' as a generic term for 
the associated emotional values. Risk weighting has
similarly both cognitive and emotional components
\cite{panno2013emotion}, where the latter are of
particular importance for long-term, viz strategic 
decision taking \cite{gilkey2010emotional}. One feels
reassured if a specific outlook is both positive and 
certain, and uncomfortable otherwise. 

The picture emerging from affective neuroscience studies
is that the brain uses deductive reasoning for the 
analysis of behavioral options and emotional states for 
the respective weighting. A larger number of distinct types 
of emotional states \cite{pfister2008multiplicity}, like 
anger, pride, fear, trust, etc, is consequently needed
when the space of accessible behavioral options increases 
\cite{schlosser2013feeling}.

\section{Cogno-emotional architectures}

A minimal precondition for application scenaria 
incorporating a basic cogno-emotional feedback loop 
is the option for the program to switch between tasks
\cite{rumbell2012emotions}. An example is a multi-gaming 
environment for which the program decides on its 
own, as detailed out further below, which game to play 
next.

\subsection{Multi-gaming environments}

We consider an architecture able to play several
games, such as Go, chess, Starcraft or console games 
like Atari. The opponents may be either human players
that are drawn from a standard internet-based matchmaking 
systems, standalone competing algorithms or agents 
participating in a multi-agent challenge 
setup \cite{samvelyan2019starcraft}. Of minor relevance
is the expertise level of the architecture and 
whether game-specific algorithms are used. A 
single generic algorithm \cite{silver2017masteringGo},
such as standard Monte Carlo tree search supplemented 
by a value and policy generating deep network
\cite{silver2017mastering}, 
would do the job. For our purpose, a key 
issue is the question whether the process 
determining which game to play is universal, in 
the sense that it can be easily adapted when the
palette of tasks is enlarged, f.i.\ when the
option to connect to a chat room is added.

For a complete cogno-emotional feedback loop 
an agent able to reason logically on an at least 
rudimentary level would be needed. This does not 
hold for the application scenario considered here.
As a consequence, one may incorporate the feedback 
of the actions of the agent onto its emotional 
states and the emotional priming of the decision
process, but not a full-fledged cognitive control 
of emotions.

\subsection{Emotional evaluation criteria}

In a first step one has to define the qualia 
of the emotional states and how they are
evaluated, viz the relation of distinct
emotions to experiences. The following
definitions are examples.
\begin{itemize}
\item[--] {\sl Satisfaction.} Winning a game raises 
	  the satisfaction level. This could hold 
          in particular for complex games, that is for 
          games that are characterized, f.i., by an
          elevated diversity of game situations.
\item[--] {\sl Challenge.} Certain game statistics may
          characterize a game as challenging. An example
          would be games for which the probability to win
          dropped temporarily precariously low.
\item[--] {\sl Boredom.} Games for which the probability
          to win remains constantly high could be classified 
          as boring or, alternatively, as relaxing. The 
          same holds for overly long games.
\end{itemize}
Emotions correspond to value-encoding variables, 
denoted here with $S$, $C$ and $B$, for 
satisfaction, challenge and boredom. Games
played are evaluated using a set of explicit 
evaluation criteria, as formulated above. An 
important note is that the aim of our framework 
is to model key functional aspects of human 
emotions, which implies that there is no need,
as a matter of principle, for the evaluation criteria 
to resemble human emotions in terms of their qualia. 
The latter is however likely to make it easier to 
develop an intuitive understanding of emotionally-driven
robotic behavior.  

\begin{figure*}[!t]
\centering
\includegraphics[width=0.75\textwidth]{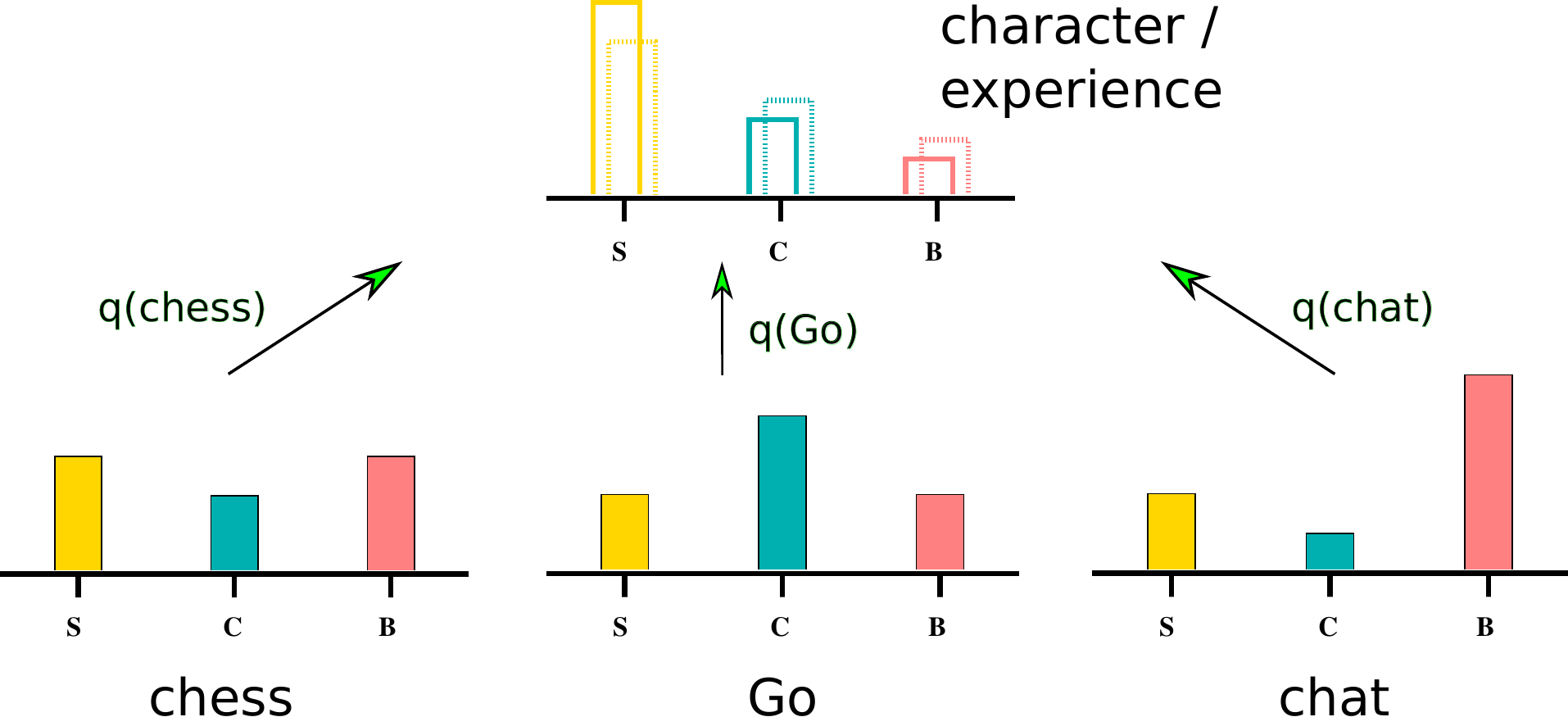}
\caption{Aligning experience with character. Behavioral
options (playing chess, playing Go, joining a chat) are 
evaluated along emotional criteria, such as being satisfying (S), 
challenging (C) or boring (B). The corresponding probability
distributions are superimposed with weights 
$q_\alpha=q(\alpha)$, where 
$\alpha\in\{\mathrm{chess},\mathrm{Go},\mathrm{chat}\}$.
See Eq.~(\ref{E_total}). The goal is to align a predefined
target distribution of emotional states, the character,
with the actual emotional experience. This can be achieved
by optimizing the probabilities $q_\alpha$ to engage in
activity $\alpha$.
}
\label{fig_characterExperience}
\end{figure*}

\subsection{Direct emotional drivings vs.\ emotional priming}

Standard approaches to modeling synthetic 
approaches often assume that emotional state
variables are explicit drivers of actions 
\cite{rodriguez2015computational}, either 
directly or via a set of internal motivations
\cite{velsquez1997modeling}. Here we are interested
in contrast in frameworks that are generic in the
sense that behavior is only indirectly influenced by
emotional states \cite{beeler2014kinder}.

In our case the agent updates in a first step 
its experience. For every type of activity, 
say when playing Go, the probability that a game 
of this type is challenging, boring or satisfying 
is continuously updated. It could be, e.g., that 
Go games are typically more challenging and less boring 
than chess games. Based on this set of data, the 
experience, the next game will be selected with 
the aim to align experience as close as possible 
with the `character' of the agent, as defined in
the following.

\subsection{Aligning experience with character}

We define the character $C_A$ of the agent as
a probability distribution of emotional 
states,
\begin{equation}
C_A=\big\{P_S,P_C,P_B\big\},
\qquad\quad
P_S+P_C+P_B=1\,,
\label{C_A}
\end{equation}
where $P_S,P_C,P_B>0$ are the target frequencies
to experience a given emotional state. Agents with 
a large $P_C$ would prefer for example challenging 
situations. The overall objective function of the agent 
is to align experience with his character.

On a basic level, experience is expressed as
a set of $N$ probability distribution functions,
\begin{equation}
E^\alpha=\big\{p_S^\alpha,p_C^\alpha,p_B^\alpha\big\},
\qquad\quad
\alpha=1,\dots,N\,,
\label{E_alpha}
\end{equation}
where $N$ is the number of possible activities
(playing Go, chess, connecting to a chat room, ...).
For every option $\alpha$ the agent records,
as illustrated in Fig.~\ref{fig_characterExperience},  
the probability $p^\alpha_i$ for the activity 
to be satisfying/challenging/boring ($i=S/C/B$).
Defining with $q_\alpha$ the likelihood 
to engage in activity $\alpha$, the overall 
experience $E_A$ is given as
\begin{equation}
E_A = \sum_\alpha q_\alpha E^\alpha,
\qquad\quad
\sum_\alpha q_\alpha =1\,,
\label{E_total}
\end{equation}
where the $E^\alpha$ are defined in (\ref{E_alpha}). 
The global objective, to align character $C_A$ and 
experience $E_A$, is achieved by minimizing the 
Kullback-Leibler divergence between $C_A$ and $E_A$ 
with respect to the $q_\alpha$. This strategy, which 
corresponds to a greedy approach, can be supplemented
by an explorative component that allows to sample new 
opportunities \cite{auer2002using}. Modulo exploration,
an activity $\alpha$ is hence selected with 
probability $q_\alpha$ 

\subsection{Stationarity principle}

Our framework is based on aligning two probability 
distribution functions, $E_A$ and $C_A$, an 
information-theoretical postulate that has been denoted 
the `stationarity principle' \cite{echeveste2015fisher}
in the context of neuronal learning \cite{trapp2018ei}.
It states, that not the activity as such should
be optimized, but the distribution of activities.
The resulting state is then varying in time, but
stationary with respect to its statistical properties.
The underlying principle of the here presented framework
corresponds to `time allocation via emotional 
stationarity' (TAES). Within this approach the 
character of the agent serves as a guiding functional,
a stochastic implementation of the principle of guided 
self-organization \cite{gros2014generating}.

\subsection{Motivational drives}

Up to now we considered purely stochastic decision 
making, namely that activities are selected probabilitistically,
as determined by the selection probabilities $q_\alpha$. An 
interesting extension are deterministic components that 
correspond to emotional drives. Considering finite time 
spans, we denote with $p_i(N_a)$ the relative number of 
times that emotion $i$ has been experienced over the course 
of the last $N_a$ activities. Ideally, the trailing averages 
$p_i(N_a)$ converge to the desired frequencies $P_i$. Substantial 
fluctuations may however occur, for example when the agent
is matched repeatedly to opponents with low levels of expertise,
which may lead to an extended series of boring games.
The resulting temporary discrepancy,
\begin{equation}
M_i = P_i-p_i(N_a)\,,
\label{M_i}
\end{equation}
between desired and trailing emotion probabilities can
then be regarded as an emotional drive. Stochastically,
$M_i$ averages out, as far as possible, when selecting
appropriate probabilities $q_\alpha$ to select an
activity $\alpha$. On a shorter time scale one may
endorse the agent with the option to reduce excessive
values of $M_k$ by direct action, viz by selecting
an activity $\beta$ characterized by large/small $p_k^\beta$
when $M_k$ is strongly positive/negative. This is however
only meaningful if the distribution $\{p_i^\beta\}$ is
peaked and not flat. Emotional drives correspond in 
this context to a additional route for reaching 
the overall goal, the alignment of experience with 
character.

\subsection{Including utility maximization}

In addition to having emotional motivations, agents
will in general be expected to maximize one or more
reward functions, like gaining credits for wining 
games or answering questions in a chat room. Without 
emotional constraints, the program would just select 
the most advantageous option, given that all options 
have already been explored in sufficient depth, in analogy 
to the multi-armed bandit problem \cite{vermorel2005multi}. 
An interesting constellation arises when rewards are weighted emotionally, 
e.g.\ with the help of the Kullback-Leibler divergence 
$D_\alpha$ between the character and the emotional experience 
of a given behavioral option \cite{gros2015complex},
\begin{equation}
D_\alpha = \sum_i P_i\log\left(\frac{P_i}{p_i^\alpha}\right)\,.
\label{D_alpha}
\end{equation}
Credits received from behavioral options that conform
with the character of the agent, having a small $D_\alpha$,
would be given a higher weight than credits gained when
engaging in activities characterized by a large $D_\alpha$.
There are then two conflicting goals, to maximize the
weighted utility and to align experience with character,
for which a suitable prioritization or Pareto optimality
may be established \cite{sener2018multi}.

Instead of treating utility as a separate feature,
one may introduce a new emotional trait, the desire 
to receive rewards, and subsume utility under
emotional optimization. Depending on the target
frequency $P_U$ to generate utility, the agent will
select its actions such that the full emotional spectrum
is taken into account. A relative weighting of
utility gains, as expressed by (\ref{D_alpha}), is
then not necessary.

\section{Discussion}

Computational models of emotions have focused
traditionally on the interconnection between emotional 
stimuli, synthetic emotions and emotional responses 
\cite{rodriguez2015computational}. A typical 
goal is to generate believable behaviors of 
autonomous social agents \cite{scherer2009emotions},
in particular in connection with psychological
theories of emotions, involving f.i.\ appraisal, 
dimensional aspects or hierarchical structures
\cite{rodriguez2015computational}. Closer to the
scope of the present investigation are proposals
that relate emotions to learning and such to behavioral
choices \cite{gadanho2003learning}. One possibility
is to use homeostatic state variables, encoding f.i.\
`well-being', for the regulation of reinforcement 
learning \cite{moerland2018emotion}. Other state 
variables could be derived from utility optimization, 
like water and energy uptake, or appraisal concepts 
\cite{moerland2018emotion}, with the latter being
examples for the abstract evaluation criteria used 
in the TAES framework. One route to measure well-being 
consist in grounding it on the relation between 
short- and long-term trailing reward rates 
\cite{broekens2007affect}. Well-being can then be used 
to modulate dynamically the balance between exploitation 
(when doing well) and exploration (when things are not 
as they used to be). Alternatively, emotional states 
may impact the policy \cite{kuremoto2013improved}.

Going beyond the main trust of research in synthetic
emotions, to facilitate human-computer interaction and
and to use emotions to improve the performance of 
machine learning algorithms that are applied to dynamic 
landscapes, the question that has been asked here 
regards how an ever ongoing sequence of distinct 
tasks can be generated by optimizing emotional 
experience, in addition to reward. Formulated as a 
time allocation problem, the rational of this 
approach is drawn mainly from affective neuroscience 
\cite{gros2009emotions}, and only to a lesser extend 
from psychological conceptualizations of 
human emotional responses. Within this setting, 
the TAES framework captures the notion that a 
central role of emotions is to serve as abstract 
evaluation tools that are to be optimized as a set, 
and not individually. This premise does not 
rule out alternative emotional functionalities.

\section{Conclusion}

Frameworks for synthetic emotions are especially 
powerful and functionally close to human emotions 
if they can be extended with ease
along two directions. First, when the protocol 
for the inclusion of new behavioral options is 
applicable to a wide range of activity classes.
This is the case when emotions do not correspond 
to specific features, but to abstract evaluation 
criteria. A given activity could then be evaluated 
as being boring, challenging, risky, demanding, 
easy, and so on. It is also desirable that the
framework allows for the straightforward inclusion 
of new traits of emotions, such as frustration.

Two agents equipped with the identical framework can 
be expected to be able to show distinct behaviors, in 
analogy to the observation that human decision making is 
generically dependent on the character of the acting 
person. For synthetic emotions this implies that there 
should exist a restricted set of parameters controlling 
the balancing of emotional states in terms of a 
preferred distribution, the functional equivalent of 
character. As realized by the TAES framework, the 
overarching objective is to adjust the relative 
frequencies to engage in a specific task, such that 
the statistics of the experienced emotional states 
aligns with the character.

Choosing between competing reward options can be done 
using a variety of strategies \cite{jordan2015machine}. 
An example is the multi-armed bandits problem, for which 
distinct behavioral options yield different rewards that
are initially not known \cite{vermorel2005multi}. Human 
life is characterized in comparison by behavioral options, 
to study, to visit a friend, to take a swim in the pool, 
and so on, that have strongly varying properties that come
with multi-variate reward dimensions. As a consequence
we proposed to define utility optimization not in terms 
of money-like credits, as it is the case for the multi-armed 
bandits problem, but on an abstract level. For this one needs 
evaluation criteria that are functionally equivalent to emotions. 
In this perspective, life-long success depends not only on the 
algorithmic capability to handle specific tasks, but also 
on the character of the agent.





\bibliographystyle{IEEEtran}

\begin{thebibliography}{10}
\providecommand{\url}[1]{#1}
\csname url@samestyle\endcsname
\providecommand{\newblock}{\relax}
\providecommand{\bibinfo}[2]{#2}
\providecommand{\BIBentrySTDinterwordspacing}{\spaceskip=0pt\relax}
\providecommand{\BIBentryALTinterwordstretchfactor}{4}
\providecommand{\BIBentryALTinterwordspacing}{\spaceskip=\fontdimen2\font plus
\BIBentryALTinterwordstretchfactor\fontdimen3\font minus
  \fontdimen4\font\relax}
\providecommand{\BIBforeignlanguage}[2]{{%
\expandafter\ifx\csname l@#1\endcsname\relax
\typeout{** WARNING: IEEEtran.bst: No hyphenation pattern has been}%
\typeout{** loaded for the language `#1'. Using the pattern for}%
\typeout{** the default language instead.}%
\else
\language=\csname l@#1\endcsname
\fi
#2}}
\providecommand{\BIBdecl}{\relax}
\BIBdecl

\bibitem{silver2017masteringGo}
D.~Silver,  \emph{et~al.}, ``Mastering the game of go without human
  knowledge,'' \emph{Nature}, vol. 550, no. 7676, pp. 354--359, 2017.

\bibitem{gao2017classification}
X.~W. Gao, R.~Hui, and Z.~Tian, ``Classification of ct brain images based on
  deep learning networks,'' \emph{Computer methods and programs in
  biomedicine}, vol. 138, pp. 49--56, 2017.

\bibitem{kumar2017opportunities}
V.~Kumar and N.~Michael, ``Opportunities and challenges with autonomous micro
  aerial vehicles,'' in \emph{Robotics Research}.\hskip 1em plus 0.5em minus
  0.4em\relax Springer, 2017, pp. 41--58.

\bibitem{malfaz2011biologically}
M.~Malfaz, {\'A}.~Castro-Gonz{\'a}lez, R.~Barber, and M.~A. Salichs, ``A
  biologically inspired architecture for an autonomous and social robot,''
  \emph{IEEE Transactions on Autonomous Mental Development}, vol.~3, no.~3, pp.
  232--246, 2011.

\bibitem{gros2013emotional}
C.~Gros, ``Emotional control--conditio sine qua non for advanced artificial
  intelligences?'' in \emph{Philosophy and Theory of Artificial
  Intelligence}.\hskip 1em plus 0.5em minus 0.4em\relax Springer, 2013, pp.
  187--198.

\bibitem{deb2014multi}
K.~Deb, ``Multi-objective optimization,'' in \emph{Search methodologies}.\hskip
  1em plus 0.5em minus 0.4em\relax Springer, 2014, pp. 403--449.

\bibitem{panksepp2004affective}
J.~Panksepp, \emph{Affective neuroscience: The foundations of human and animal
  emotions}.\hskip 1em plus 0.5em minus 0.4em\relax Oxford university press,
  2004.

\bibitem{gros2010cognition}
C.~Gros, ``Cognition and emotion: perspectives of a closing gap,''
  \emph{Cognitive Computation}, vol.~2, no.~2, pp. 78--85, 2010.

\bibitem{baumeister2007emotion}
R.~F. Baumeister, K.~D. Vohs, C.~Nathan~DeWall, and L.~Zhang, ``How emotion
  shapes behavior: Feedback, anticipation, and reflection, rather than direct
  causation,'' \emph{Personality and social psychology review}, vol.~11, no.~2,
  pp. 167--203, 2007.

\bibitem{beeler2014kinder}
J.~A. Beeler, R.~Cools, M.~Luciana, S.~B. Ostlund, and G.~Petzinger, ``A
  kinder, gentler dopamine... highlighting dopamine's role in behavioral
  flexibility,'' \emph{Frontiers in neuroscience}, vol.~8, 2014.

\bibitem{ochsner2005cognitive}
K.~N. Ochsner and J.~J. Gross, ``The cognitive control of emotion,''
  \emph{Trends in cognitive sciences}, vol.~9, no.~5, pp. 242--249, 2005.

\bibitem{inzlicht2015emotional}
M.~Inzlicht, B.~D. Bartholow, and J.~B. Hirsh, ``Emotional foundations of
  cognitive control,'' \emph{Trends in cognitive sciences}, vol.~19, no.~3, pp.
  126--132, 2015.

\bibitem{cutuli2014cognitive}
D.~Cutuli, ``Cognitive reappraisal and expressive suppression strategies role
  in the emotion regulation: an overview on their modulatory effects and neural
  correlates,'' \emph{Frontiers in Systems Neuroscience}, vol.~8, 2014.

\bibitem{miller2018happily}
M.~Miller and A.~Clark, ``Happily entangled: prediction, emotion, and the
  embodied mind,'' \emph{Synthese}, vol. 195, no.~6, pp. 2559--2575, 2018.

\bibitem{pessoa2008relationship}
L.~Pessoa, ``On the relationship between emotion and cognition,'' \emph{Nature
  reviews neuroscience}, vol.~9, no.~2, p. 148, 2008.

\bibitem{pessoa2019embracing}
------, ``Embracing integration and complexity: placing emotion within a
  science of brain and behaviour,'' \emph{Cognition and Emotion}, vol.~33,
  no.~1, pp. 55--60, 2019.

\bibitem{pessoa2018understanding}
------, ``Understanding emotion with brain networks,'' \emph{Current opinion in
  behavioral sciences}, vol.~19, pp. 19--25, 2018.

\bibitem{phelps2006emotion}
E.~A. Phelps, ``Emotion and cognition: insights from studies of the human
  amygdala,'' \emph{Annu. Rev. Psychol.}, vol.~57, pp. 27--53, 2006.

\bibitem{morawetz2016neural}
C.~Morawetz, S.~Bode, J.~Baudewig, A.~M. Jacobs, and H.~R. Heekeren, ``Neural
  representation of emotion regulation goals,'' \emph{Human brain mapping},
  vol.~37, no.~2, pp. 600--620, 2016.

\bibitem{lerner2015emotion}
J.~S. Lerner, Y.~Li, P.~Valdesolo, and K.~S. Kassam, ``Emotion and decision
  making,'' \emph{Annual review of psychology}, vol.~66, pp. 799--823, 2015.

\bibitem{reimann2010somatic}
M.~Reimann and A.~Bechara, ``The somatic marker framework as a neurological
  theory of decision-making: Review, conceptual comparisons, and future
  neuroeconomics research,'' \emph{Journal of Economic Psychology}, vol.~31,
  no.~5, pp. 767--776, 2010.

\bibitem{panno2013emotion}
A.~Panno, M.~Lauriola, and B.~Figner, ``Emotion regulation and risk taking:
  Predicting risky choice in deliberative decision making,'' \emph{Cognition \&
  emotion}, vol.~27, no.~2, pp. 326--334, 2013.

\bibitem{gilkey2010emotional}
R.~Gilkey, R.~Caceda, and C.~Kilts, ``When emotional reasoning trumps iq,''
  \emph{harvard business review}, vol.~88, no.~9, p.~27, 2010.

\bibitem{pfister2008multiplicity}
H.-R. Pfister and G.~B{\"o}hm, ``The multiplicity of emotions: A framework of
  emotional functions in decision making,'' \emph{Judgment and decision
  making}, vol.~3, no.~1, p.~5, 2008.

\bibitem{schlosser2013feeling}
T.~Schl{\"o}sser, D.~Dunning, and D.~Fetchenhauer, ``What a feeling: the role
  of immediate and anticipated emotions in risky decisions,'' \emph{Journal of
  Behavioral Decision Making}, vol.~26, no.~1, pp. 13--30, 2013.

\bibitem{rumbell2012emotions}
T.~Rumbell, J.~Barnden, S.~Denham, and T.~Wennekers, ``Emotions in autonomous
  agents: comparative analysis of mechanisms and functions,'' \emph{Autonomous
  Agents and Multi-Agent Systems}, vol.~25, no.~1, pp. 1--45, 2012.

\bibitem{samvelyan2019starcraft}
M.~Samvelyan, T.~Rashid, C.~S. de~Witt, G.~Farquhar, N.~Nardelli, T.~G. Rudner,
  C.-M. Hung, P.~H. Torr, J.~Foerster, and S.~Whiteson, ``The starcraft
  multi-agent challenge,'' \emph{arXiv preprint arXiv:1902.04043}, 2019.

\bibitem{silver2017mastering}
D.~Silver, T.~Hubert, J.~Schrittwieser, I.~Antonoglou, M.~Lai, A.~Guez,
  M.~Lanctot, L.~Sifre, D.~Kumaran, T.~Graepel \emph{et~al.}, ``Mastering chess
  and shogi by self-play with a general reinforcement learning algorithm,''
  \emph{arXiv preprint arXiv:1712.01815}, 2017.

\bibitem{rodriguez2015computational}
L.-F. Rodr{\'\i}guez and F.~Ramos, ``Computational models of emotions for
  autonomous agents: major challenges,'' \emph{Artificial Intelligence Review},
  vol.~43, no.~3, pp. 437--465, 2015.

\bibitem{velsquez1997modeling}
J.~Velsquez, ``Modeling emotions and other motivations in synthetic agents,''
  in \emph{Proc. 14th Nat. Conf. Artif. Intell}, 1997, pp. 10--15.

\bibitem{auer2002using}
P.~Auer, ``Using confidence bounds for exploitation-exploration trade-offs,''
  \emph{Journal of Machine Learning Research}, vol.~3, no. Nov, pp. 397--422,
  2002.

\bibitem{echeveste2015fisher}
R.~Echeveste, S.~Eckmann, and C.~Gros, ``The fisher information as a neural
  guiding principle for independent component analysis,'' \emph{Entropy},
  vol.~17, no.~6, pp. 3838--3856, 2015.

\bibitem{trapp2018ei}
P.~Trapp, R.~Echeveste, and C.~Gros, ``Ei balance emerges naturally from
  continuous hebbian learning in autonomous neural networks,'' \emph{Scientific
  reports}, vol.~8, no.~1, p. 8939, 2018.

\bibitem{gros2014generating}
C.~Gros, ``Generating functionals for guided self-organization,'' \emph{Guided
  Self-Organization: Inception}, pp. 53--66, 2014.

\bibitem{vermorel2005multi}
J.~Vermorel and M.~Mohri, ``Multi-armed bandit algorithms and empirical
  evaluation,'' in \emph{European conference on machine learning}.\hskip 1em
  plus 0.5em minus 0.4em\relax Springer, 2005, pp. 437--448.

\bibitem{gros2015complex}
C.~Gros, \emph{Complex and adaptive dynamical systems: A primer}.\hskip 1em
  plus 0.5em minus 0.4em\relax Springer, 2015.

\bibitem{sener2018multi}
O.~Sener and V.~Koltun, ``Multi-task learning as multi-objective
  optimization,'' in \emph{Advances in Neural Information Processing Systems},
  2018, pp. 527--538.

\bibitem{scherer2009emotions}
K.~R. Scherer, ``Emotions are emergent processes: they require a dynamic
  computational architecture,'' \emph{Philosophical Transactions of the Royal
  Society B: Biological Sciences}, vol. 364, no. 1535, pp. 3459--3474, 2009.

\bibitem{gadanho2003learning}
S.~C. Gadanho, ``Learning behavior-selection by emotions and cognition in a
  multi-goal robot task,'' \emph{Journal of Machine Learning Research}, vol.~4,
  no. Jul, pp. 385--412, 2003.

\bibitem{moerland2018emotion}
T.~M. Moerland, J.~Broekens, and C.~M. Jonker, ``Emotion in reinforcement
  learning agents and robots: a survey,'' \emph{Machine Learning}, vol. 107,
  no.~2, pp. 443--480, 2018.

\bibitem{broekens2007affect}
J.~Broekens, W.~A. Kosters, and F.~J. Verbeek, ``On affect and self-adaptation:
  Potential benefits of valence-controlled action-selection,'' in
  \emph{International Work-Conference on the Interplay Between Natural and
  Artificial Computation}.\hskip 1em plus 0.5em minus 0.4em\relax Springer,
  2007, pp. 357--366.

\bibitem{kuremoto2013improved}
T.~Kuremoto, T.~Tsurusaki, K.~Kobayashi, S.~Mabu, and M.~Obayashi, ``An
  improved reinforcement learning system using affective factors,''
  \emph{Robotics}, vol.~2, no.~3, pp. 149--164, 2013.

\newpage

\bibitem{gros2009emotions}
C.~Gros, ``Emotions, diffusive emotional control and the motivational problem
  for autonomous cognitive systems,'' in \emph{Handbook of Research on
  Synthetic Emotions and Sociable Robotics: New Applications in Affective
  Computing and Artificial Intelligence}.\hskip 1em plus 0.5em minus
  0.4em\relax IGI Global, 2009, pp. 119--132.

\bibitem{jordan2015machine}
M.~I. Jordan and T.~M. Mitchell, ``Machine learning: Trends, perspectives, and
  prospects,'' \emph{Science}, vol. 349, no. 6245, pp. 255--260, 2015.

\end{thebibliography}

\end{document}